# Using Interval Particle Filtering for Marker less 3D Human Motion Capture

Jamal Saboune
INRIA-LORIA, B.P.239
54506 Vandoeuvre Lès Nancy, France
saboune@loria.fr

François Charpillet
INRIA-LORIA, B.P.239
54506 Vandoeuvre Lès Nancy, France
charpillet@loria.fr

## Abstract

*In this paper we present a new approach for marker less human motion capture from conventional camera feeds. The aim of our study is to recover 3D positions of key points of the body that can serve for gait analysis. Our approach is based on foreground segmentation, an articulated body model and particle filters. In order to be generic and simple no restrictive dynamic modelling was used. A new modified particle filtering algorithm was introduced. It is used efficiently to search the model configuration space. This new algorithm which we call Interval Particle Filtering reorganizes the configurations search space in an optimal deterministic way and proved to be efficient in tracking natural human movement. Results for human motion capture from a single camera are presented and compared to results obtained from a marker based system. The system proved to be able to track motion successfully even in partial occlusions.*

## 1. Medical framework

Recent studies showed that falls have become the first cause of accidental mortality among the elderly, which reveals a very serious problem in a population with growing age average. Most of the falls occur during walking. The aim of our study is to propose a methodology and a technology to detect a tendency towards the fall of seniors, observing their everyday activities at home. Gait parameters would be evaluated to calculate a dynamic balance indicator. The originality of our approach implicates many constraints to the methods and technologies used. Actually, the senior living environment should not be altered and no wearable sensors will be attached to the body. Our system must be capable of evaluating the balance automatically without any human intervention. On the other hand, used sensors should be low cost. In these conditions, using video feeds seems to be the most adequate approach to measure gait parameters. In this paper we will present a new method to track body 3D motion, which respects the principles of our approach.

## 2. Human motion capture

An in-depth gait analysis requires the knowledge of elementary spatio-temporal parameters such as walking speed, hip and knee angles, stride length and width, time of support, among others. In order to obtain this information, a 3D human motion capture system has to be developed. Marker-based systems [1] have been widely used for years with applications found in biometrics. In typical systems, a number of reflective markers are attached to several key points of the patient's body and then captured by infrared cameras fixed at known positions in the footage environment. The markers positions are then transformed into 3D positions using triangulation from the several cameras feeds, making it impossible to track a point's motion when it is not visible by two or more cameras. However, using markers could be considered obtrusive. It also implicates the use of expensive specialized equipment and requires a footage taken in a specially arranged environment. Using video feeds from conventional cameras and without the use of special hardware, implicates the development of a marker less body motion capture system. Research in this domain is generally based on the articulated-models approach. Haritaoglu et al. [2] present an efficient system capable of tracking 2D body motion using a single camera. This might be used in many applications. However it is unable to provide 3D positions, restricting the information we can extract from the feeds. Bregler et al. [3] used gradient descent search with frame-to-frame region based matching and applied this method on short multi camera sequences. This method proved to be unable to track agile motions with cluttered backgrounds. On the other hand, locating body parts by matching image regions, risks to produce a drift in long sequences. Combining 2D tracker and learned 3D configuration models, Howe et al. [4] were able to produce 3D body pose from short single camera feeds. Gavrila and Davis [5] use an explicit hierarchical search, in which they sequentially locate parts of the body's *kinematic* chain (e.g. torso, followed by upper arm, lower arm and then hand), reducing the search complexity significantly. In real world situations, it seems to be hard to specify each body part in the image independently without using labels or colour cues. Sidenbladh et al. [6] use Condensation algorithm [7] with learned stochastic models and a generative model of image formation to track full body motion. The large number of particles used, makes this algorithm run slowly. Cohen et al. [8] tried to reduce the number of particles, using Support Vector Machine, to train body models. However, using dynamic models would restrict the generality of the approach and prevent the system from tracking gait abnormalities. Using multiple cameras feeds at a 60 frames/s capture, Deutscher et al. [9] produced the best known results to date in 3D full body tracking. Their approach was based on weak dynamical modelling and on *annealed particle filtering*, which is a complex modified Condensation algorithm. In fact, a multi layered particle based stochastic search algorithm was applied to reduce the number of particles. This algorithm uses a continuation principle to introduce the influence of narrow peaks in the *weight* function gradually. Applying this type of layered search augments the risks of falling into local minima, especially in the case of a lower frequency capture, and did not prove to reduce the time complexity in a significant way.

Most of these methods were originally developed and used for character animation and do not meet the requirements of our study. In fact we need to extract the exact 3D position of several points of the human body in order to detect gait abnormalities, using conventional digital camera feeds (25 frames/s) only. Respecting these conditions requires conceiving a new simple algorithm. The method we present here is based on a simple modified particle filtering algorithm, which we call *Interval Particle Filtering (IPF)*. Image foreground segmentation and 3D articulated body modelling are basic elements in our approach. The articulated model and the likelihood function we use will be described in Sections 3 and 4. A review of particle filters will be exposed in Section 5. In section 6 we present the *Interval Particle Filtering* algorithm, results will be shown in Section 7.

## 3. The articulated body model

As in most full body tracking approaches, an articulated body model was used. This model is formed by 19 points or joints that represent key points of the human body (head, elbows, sacrum, knees, ankles etc.). 17 rigid segments then join up these points. Each point is given a number of degrees of freedom (3 maximum), representing the rotations about 3D axes (x, y, z).

These degrees of freedom should simulate approximately the way the human body parts move. Our model is parameterised with 31 degrees of freedom. This skeleton model is then fleshed out in a way to have a visually realistic body representation (Figure 1). For each freedom degree, we define a range beyond which no movement is allowed. These constraints can easily be modified depending on the nature of the actions to be tracked. For example, in a standard walking situation, the leg's rotation cannot take values greater than 60 degrees nor lesser than -30 degrees. Due to the nature of the human body, our model is composed of *kinematic* chains; a body part whose movement implicates the movement of another body part, forms a *kinematic* chain with the latter.

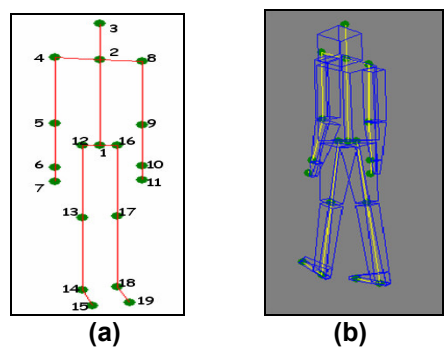

**Figure 1. Articulated Model; (a) The skeleton articulated model defined by 19 key points of the human body and 17 segments. This model is composed of 4 *kinematic* chains. Each point is given a number of freedom degrees (rotations). Due to restrictions in human body parts motion this model can be represented by a 31degrees of freedom vector. The model is fleshed out by adding volumes around the segments.**

Deutscher et al. [9] combined a similar model with a dynamic modelling approach; in fact, they use a velocity model for each joint's motion from the previous frame, in order to predict the body pose in the next frame, which restricts the capacity to detect sudden changes in movement if the frames are distant in time. Sidenbladh et al. [6] introduced trained dynamical models. Trained models are of great interest for robust tracking, but they force the real motions tracked to be similar to those observed in the training set. The use of these models would make it unable to detect the abnormal movements we are interested in, due to the fact that these movements would not necessarily be in the training set. In addition it seems impractical to pre-train models for each possible situation and movement of the body. The use of trained models being in contradiction with the goal of our study, we opted not to use any. No dynamic model was used neither, which makes our approach simple and generic.

## 4. The Likelihood function

In addition to defining a model, a method to evaluate how well its configuration fits with the image should be developed. Our model is configured by a set of values of the 31 degrees of freedom. The likelihood function of this configuration, representing the degree of similarity with the image data, is called *weight* in a particle filtering context. In marker-based systems, the markers positions in each camera's image plan give us the real 3D positions of the markers. In some approaches [6] [8], texture mapping was used to realistically render body images. Despite the advantages it provides, creating this type of images would be specific for each person and the conditions of the video capture (light, clothes etc.). Edge detection and foreground segmentation were used to construct a simple and general *weight* function in [9]. We choose to use a simple foreground segmentation to construct our *weighting* function. Actually, we construct the *silhouette* image by subtracting pixel by pixel the background from the current image and then applying a threshold filter. This image will then be compared to the synthetic image representing our model's configuration (2D projection of our 3D body's model) to which we want to assign the *weight*. After subtracting the synthetic image from the silhouette, the weight function will be calculated by:

$$w = \frac{N_c}{N_s + N_m}$$

where:
$N_c$ = Number of common pixels between the *silhouette* and the synthetic image
$N_s$ = Number of pixels representing the *silhouette* - $N_c$
$N_m$ = Number of pixels of the synthetic image representing the model - $N_c$

The choice of this function is motivated by the fact that we aim to find the model's configuration that maximises the likelihood of its 2D projection to the real body pose. This can be interpreted by a higher $N_c$ and a lower ($N_s + N_m$). In case of multiple cameras, the weight function $w$ would be:

$$w = \frac{\sum_{i=1}^{c} w_i}{c}$$

where $c$ is the number of cameras, and $w_i$ the weight deducted from the image of camera $i$.

This method is simple and can be applied in any condition; however, it may present a little weakness in presence of heavy shadows and in the case of a person with loose clothing. This problem can be solved by adjusting the threshold filter.

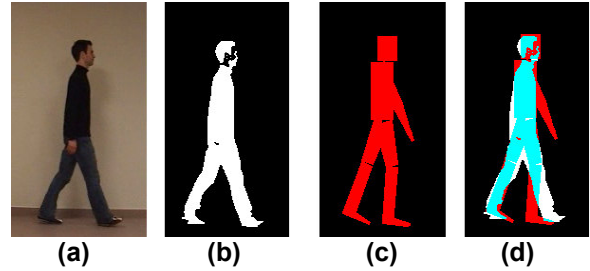

(a)     (b)     (c)     (d)

**Figure 2. Estimating the likelihood; The input image (a) will be transformed to a *silhouette* image (b) by subtracting pixel by pixel the background image and then applying a threshold filter. (c) is the synthetic 2D image representing one of the 3D model's configuration. In (d) we compare the synthetic image to the *silhouette* by subtracting the first from the second and evaluating the number $N_c$ of common pixels, the number of pixels of the silhouette outside the synthetic image $N_s$ and the number $N_m$ of pixels of the synthetic image outside the silhouette.**

## 5. Particle filters

Full body motion tracking can be treated as a Bayesian state estimation problem. Our 3D model pose is established through the configuration of the degrees of freedom values, this configuration can thus represent the state vector of the model **X**. In addition to the state model, we define an observation **Z**, through which the likelihood of a state vector at $t=t_k$ is evaluated by calculating $P(\mathbf{Z_k}/\mathbf{X})$. In our approach, this observation would be the image of the person we are tracking and the normalized *weight* of a model's configuration $m$, would represent the observation probability ($w^{(k)}(m) = P(\mathbf{Z_k}/\mathbf{X} = m)$).

The most commonly known Bayesian tracker is the Kalman Filter. In order to apply a Kalman filter, the process we observe should have a Gaussian observation probability and a Gaussian posterior probability density $P(\mathbf{X}/\underline{\mathbf{Z}}_\mathbf{k})$, where $\underline{\mathbf{Z}}_\mathbf{k}$ is the observations history at $t_k$. Deutscher [10] proved that the posterior density in human motion capture is non

Gaussian and multi-modal, thus a Kalman Filter may act poorly in this type of application. Particle filtering, also known as the Condensation algorithm [7], proved to be able to handle such type of non-Gaussian, multi-modal densities. In fact it can model uncertainty by transmitting less fitting state configurations at $t_k$, to later time steps, and thus giving them a chance to be chosen. In a particle filtering framework, each 3D model's configuration (31 degrees of freedom vector) is represented by a particle. For each particle a weight is assigned (as described in Section 4), at each time step. The posterior density is represented by the set of particles and their weights (Figure 3). A particle filter consists of 3 essential steps (at $t_k$):

- *Selection*: Among the set of N particles created at $t_{k-1}$, N particles are picked up with replacement; the probability of selecting a particle $p_m$ is proportional to its normalized weight $w^{(k-1)}(p_m)$.

- *Prediction*: Each particle is updated according to the evolution model of the system; $X_k = A_k * X_{k-1} + B_k$ ($B_k$ is a white Gaussian noise).

- *Measure*: Given the observation $Z_k$ (image at $t_k$), the *weight* for each particle is calculated. The weights are then normalized and the new *weighed* particles set is used at $t_{k+1}$. The state of the system at $t_k$ is estimated by the particle having the greatest *weight*, or the average particle of the *weighed* particles set.

The particle filtering can be viewed as a search for the best particle in a well defined particle set at each time step. In order to have a fair representation of the posterior density and a realistic state vector estimation, a certain number of particles are necessary. In a high dimensional space this number becomes relatively big. The use of a greater number of particles leads to better results. On the other hand, using more particles augments the temporal complexity of the algorithm, due to the fact that at each time step the *weight* for all particles must be calculated. The goal of all modified particle filtering algorithms is to reduce the number of particles needed, and this especially in high dimensional spaces, where the complexity could make the basic particle filtering algorithm practically inapplicable. Despite the high dimension of our state vector (31 degrees of freedom), we opted to use a particle filtering algorithm, due to its capacity to handle the multi modal observation probability. In parallel, we modified the basic algorithm by introducing the *Interval Particle Filtering* that tends to reconfigure the particles search space in an optimal way.

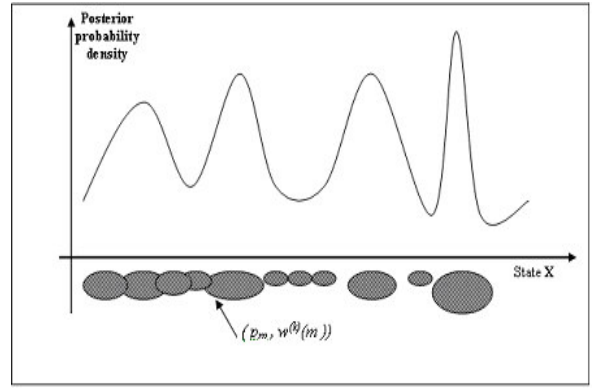

**Figure 3. The posterior probability density P(X/ $\underline{Z}_k$) is represented by the set of *weighed* particles. Each particle $p_m$ is represented by an ellipse whose surface is proportional to its weight $w^{(k)}(p_m)$.**

## 6. Interval Particle Filtering

The *Interval Particle Filtering (IPF)* introduces simple modifications on the Condensation algorithm in order to optimise the particles search space configuration. These modifications are done in a way to preserve the advantages a particle filter offers. Neither dynamic modelling nor an evolution model was used. A single iteration per time step is accomplished, excluding any layered search. We preserve the 3 steps structure of the basic particle filtering algorithm at each time step $t_k$ (Figure 4):

- *Selection*: The N particles set created at $t_{k-1}$ is sorted by its weights. In this sorted set, a number of *M* distinct particles, that have the biggest *weights*, are selected. This way, the capacity of our filter to handle multi modal densities is preserved.

- *Prediction*: As we are especially interested in some specific elements of the state vector, we can dissociate it into 2 state vectors: the first one **L** containing the 'interesting' degrees of freedom and the second **R** covering the rest of the freedom degrees. Each particle is now represented by 2 vectors instead of one. **L** is then updated and replaced by a multidimensional Interval (space) composed of *I* vectors covering a grid of vectors surrounding the initial vector **L**. If **L** has *n* dimensions, we define for each dimension $l_i$ *(i=1..n)* an interval surrounding its value. This interval is then discretised into *q* values that represent the possible values of $l_i$ at $t_k$. In order to cover all combinations of possible values of the *n* dimensions of **L**, $q^n$ vectors will be needed ($I=q^n$). This approach is inspired by the presence of physiological restrictions on the degrees of freedom evolution (e.g. maximum angular velocity of joints in human motion). As a result, each particle will be updated and replaced by *I* 'neighbour' particles so as to cover the whole possible configuration space of the 'interesting' dimensions, closely surrounding the particle's 'interesting' dimensions configuration. For each particle, **R** is then updated by adding a white noise vector. The total number N of particles will then be equal to *M\*I*. The width of the interval and the number *I* of vectors depend on the nature of the system. A wider interval and a greater *I* provide more accurate results but result in greater computational cost.

- *Measure*: This step remains unaltered; given the observation $Z_k$ (image at $t_k$), the *weight* for each particle is calculated (as described in Section 4) and the new *weighed* particles set is propagated to be used at $t_{k+1}$. The estimated state vector (body pose) at $t_k$ will be represented by the particle (the model's configuration) having the greatest *weight*.

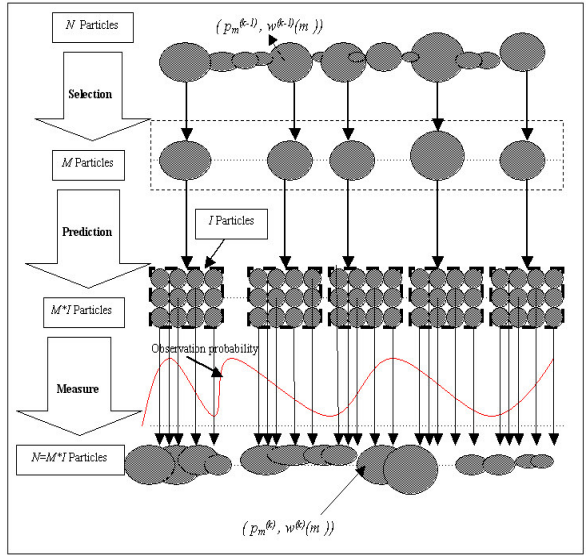

**Figure 4. IPF; One time step ($t_{k-1}$, tk) of *Interval Particle Filtering***

The Condensation algorithm used without any evolution model, will divide the set of N particles into many sub-sets of different sizes (proportional to each selected particle *weight*); the particles in each sub-set are distributed around one of the heaviest particles in a random manner. In presence of restrictions on the state vector evolution, *Interval Particle Filtering* reorganises the set of N particles into *M* sub-sets each formed of *I* particles covering in a deterministic way the 'neighbourhood' of the *j* heaviest particle *(j=1..M)* at the previous time step. Thus, the *Interval Particle Filtering* creates a fitter and more reliable search space than that created by the condensation algorithm.

## 7. Applications and results

The articulated model we use is characterised by a particular point (representing the sacrum point) which we call the body origin. Prior to estimating the body parts configuration, we should estimate the body origin's 3D position and orientation. Using the configuration of the body parts that was estimated at $t_{k-1}$ and the *silhouette* image at $t_k$ we can find the origin's 3D position and orientation. This can be done by applying hierarchical search over all possible values of this position (each dimension separately). Each of these positions produces a synthetic 2D image of the articulated model, which we compare to the *silhouette* image; the position producing the image that fits the most this *silhouette* image is considered to be the body origin's position. Knowing this position, we can apply the *Interval Particle Filtering* to estimate the body parts positions. As we are especially interested in the lower limbs movement for gait analysis, we chose to apply the algorithm with **L** representing 4 (*n*=4) of the lower limbs freedom degrees (knees and hips rotations).

As we use video feeds captured at 20 Hz, and due to the physiological constraints, we can define the interval width for each angle to be 10°. This means that the angle between two time steps (50 ms) can not evolve (positively or negatively) for more than 5°. If we discretise this interval into *q*=3 values, we get $I=3^4$ = 81 particles per interval. With *M*=81 our algorithm will be running with 6561 particles. At each time step we can get the estimated 3D positions of the 19 points forming the articulated model and the estimated values of the 31 freedom degrees. The initialisation of the body parts configuration is done automatically. In fact an exhaustive search is applied to the initial set of particles in order to find the *heaviest* particle. This initial set contains N vectors configured as to cover a well defined grid of plausible configurations.

We present results from three video captures of three different subjects in normal clothes, moving in a normal environment. Those feeds were captured at 20 frames/s using a single commercial digital video camera. Image processing was done offline, using a P4 3GHz PC. It takes 20 seconds per frame (7 minutes per second of footage) to find the body origin 3D position and its parts configuration using *Interval Particle Filtering*. Despite being too far from a real time performance, our system runs faster than many other systems developed in the literature.

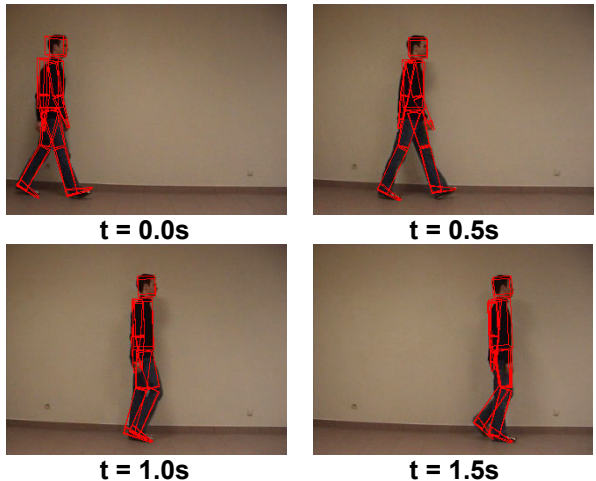

**Figure 5. Results for a subject walking normally along a wall; despite the fact that the left leg is occluded in these frames, the algorithm was able to track the movement with success.**

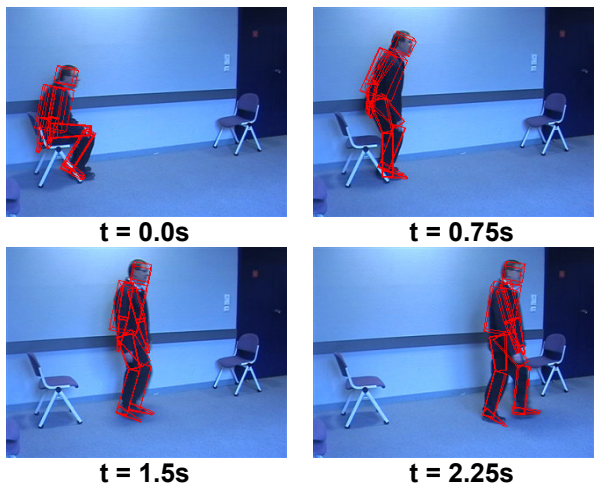

**Figure 6. Results showing a subject getting up from a chair, turning and then walking. This set of images illustrates the use of the body origin positioning algorithm that was developed. The challenge in this feed is to determine the body origin position for the person while getting up and beginning to walk.**

The first set of images (Figure 5) shows a person walking in a straight line, with some momentary occlusions. The second set (Figure 6) shows a person getting up from a chair and then turning and walking. The difficulty here is to track the movement of the person while getting up, but the algorithm succeeded in it. The third set (Figure 7) shows a person walking randomly. This scene had also been filmed and treated by a Vicon system (marker based motion capture system) running at 100 Hz. Positions of some body points produced by our algorithm can be compared favourably to the same positions produced by the Vicon, as shown in Figures 8.

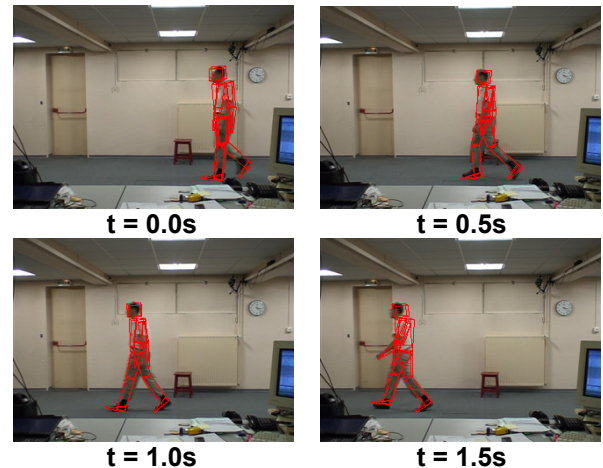

**Figure 7. Results showing a subject walking randomly. This scene had been also captured by a Vicon system composed of 6 infra red cameras. Reflective markers were fixed to the subject's body key points (the same key points of our skeleton model) and the results were compared to the results obtained using IPF algorithm (Figures 8).**

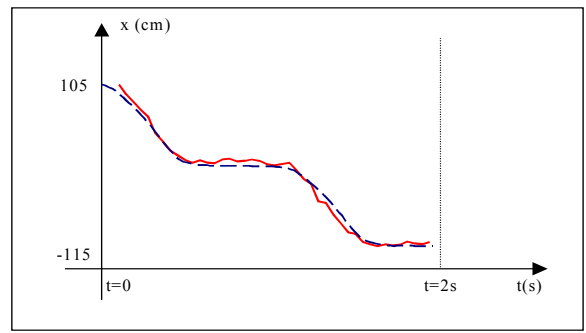

**Figure 8.1 Longitudinal movement x of the left ankle (for the subject appearing in Figure 7), estimated by IPF (in plain) and using the Vicon system (in dotted). The two curves take a similar form, however the dotted curve is smoother. This can be explained by the differences in capture and treatment frequency (20 Hz for IPF, 100 Hz for Vicon). The x values presented are calculated in the camera referential**

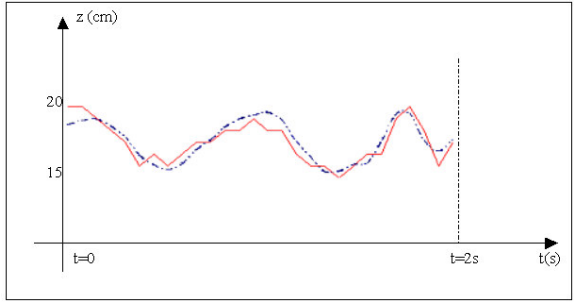

**Figure 8.2 Comparison of the estimated (by IPF) vertical movement of the sacrum point (in plain) and the same variable evaluated by Vicon (in dotted). As for Figure 8.1, the dotted curve is smoother. The values presented are calculated in the camera referential**

The fact of using a single camera prevents us from seeing some body parts (which depends on the view angle) for a long portion of the footage and the movement of these parts could thus not be evaluated fairly. Using multiple cameras would solve this problem but implicates the complexity of calibrating and adjusting a stereo vision system.

## 8. Conclusion

In this paper we presented a new approach for marker less human motion capture from a single commercial video camera, based on a modified particle filtering algorithm. The aim of our study is to extract gait parameters from the feeds, and the results obtained compared to those of a marker based system were encouraging. Using multiple cameras would improve the results obtained but at the same time complicates the method. The *Interval Particle Filtering* we introduced here, proved to give good results despite the high dimensionality of the state vector, even with occlusions. This algorithm is simple to implement and works with video feeds captured at any frequency and in any environment. Since we do not use any restrictive dynamic model, our approach can be also described as being generic.

**Acknowledgements:** This work was financed by the RNTS project *PARAChute* and the INRIA action *DialHemo*.

## 9. References

[1] VICON web based literature, URL http://www.metrics.co.uk

[2] HARITAOGLU, I., HARWOOD, D., DAVIS, L. (1998): 'w 4s: A real-time system for detecting and tracking people in 2.5D', Proc. 5th European Conf. Computer Vision 1998, Freiburg Germany, **1**, pp. 877-892

[3] BREGLER, C., MALIK, J. (1998): 'Tracking people with twists and exponential maps', Proc. Conf. Computer Vision and Pattern Recognition 1998.

[4] HOWE, N.R., LEVENTON, M.E., FREEMAN, W.T. (2000): 'Bayesian reconstruction of 3D human motion from single camera video', Advances in Neural Information Processing Systems MIT Press 2000, **12**, pp. 820-826

[5] GAVRILA, D., DAVIS, L.S. (1996): '3d model based tracking of humans in action a multi view approach', Proc. Conf. Computer Vision and Pattern Recognition 1996, pp. 73-80

[6] SIDENBLADH, H., BLACK, M.J., FLEET, D.J. (2000): 'Stochastic tracking of 3D human figures using 2d image motion', Proc. 6th European Conference on Computer Vision 2000, Dublin

[7] ISARD, M.A., BLAKE, A. (1998): 'CONDENSATION- conditional density propagation for visual tracking', International Journal of Computer Vision 1998, **29(1)**, pp. 5-28

[8] COHEN, I., MUN WAI, L. (2002): '3D Body Reconstruction for Immersive Interaction', Second international Workshop on Articulated Motion and Deformable Objects 2002, Palma de Mallorca Spain, pp. 119-130

[9] DEUTSCHER, J., BLAKE, A., REID, I. (2000): 'Articulated body motion capture by annealed particle filtering', Proc. Conf. Computer Vision and Pattern Recognition 2000, **2**, pp. 1144-1149

[10] DEUTSCHER, J., BLAKE, A., NORTH, B., BASCLE, B. (1999): 'Tracking through singularities and discontinuities by random sampling', Proc. 7th Int. Conf. Computer Vision 1999, pp. 1144-1149